\RequirePackage{fix-cm}
\documentclass[smallextended]{svjour3}       % onecolumn (second format)
\smartqed  % flush right qed marks, e.g. at end of proof
\usepackage[T1]{fontenc}
\usepackage[utf8]{inputenc}
\usepackage[final]{microtype}
\usepackage{graphicx}
\usepackage[citetracker=true,citestyle=authoryear,dashed=false,%
            bibstyle=authoryear,maxcitenames=2,maxbibnames=99,minnames=1,%
            backend=biber,uniquename=false,uniquelist=false,natbib,
            useprefix=true]{biblatex}
\usepackage{booktabs}
\usepackage[round-mode=places,
            round-precision=1,
            table-text-alignment=right,
            table-align-text-pre = false,
            table-align-text-post = false,
            table-align-comparator = false,
            table-figures-uncertainty=0,
            table-number-alignment=right,
            table-align-uncertainty=false,
            table-align-exponent=false,
            group-minimum-digits=4,
]{siunitx}
\DeclareSIUnit{\nothing}{\relax}
\usepackage{xcolor}
\usepackage{hyperref}
\hypersetup{
  breaklinks,
  bookmarksopen,bookmarksdepth=3,
  colorlinks=true,
  linkcolor=[rgb]{0 0 0.1},
  urlcolor=[rgb]{0 0 0.1},
  filecolor=[rgb]{0 0 0.1},
  citecolor=[rgb]{0 0 0.1},
  pdftitle={A survey of Turkish corpora and lexical resources}
}

\usepackage{tcolorbox}
\usepackage{orcidlink}

\newcommand{\topic}[1]{#1}
\newcommand{\weblink}[2]{\href{#1}{\color{blue!50!black}#2}{}}

\journalname{Language Resources and Evaluation}

\begin{document}

\title{Resources for Turkish Natural Language Processing}
\subtitle{A critical survey}

\author{{\c C}a{\u g}r{\i} {\c C}{\"o}ltekin  \and
        A. Seza Do{\u g}ru{\"o}z   \and
        {\"O}zlem {\c C}etino{\u g}lu
}

\authorrunning{{\c C}. {\c C}{\"o}ltekin, A. S. Do{\u g}ru{\"o}z, {\"O}. {\c C}etino{\u g}lu} % if too long for running head

\institute{{\c C}a{\u g}r{\i} {\c C}{\"o}ltekin$^{\orcidlink{0000-0003-1031-6327}}$
                \at University of T{\"u}bingen, T{\"u}bingen, Germany.
                \email{ccoltekin@sfs.uni-tuebingen.de} \and
           A. Seza Do{\u g}ru{\"o}z $^{\orcidlink{0000-0003-2589-5894}}$ \at Ghent University, Ghent, Belgium.
         \email{as.dogruoz@ugent.be} \and
           {\"O}zlem {\c C}etino{\u g}lu$^{\orcidlink{0000-0001-6100-4839}}$
                \at University of Stuttgart, Stuttgart, Germany.
                \email{ozlem@ims.uni-stuttgart.de}
}

\date{Received: date / Accepted: date}

\maketitle

\begin{tcolorbox}[colback=white]
  The final version of this paper is published at \url{https://doi.org/10.1007/s10579-022-09605-4}.
  Please cite the published version.
\end{tcolorbox}
\begin{abstract}
  This paper presents a comprehensive survey of corpora and lexical resources
  available for Turkish.
  We review a broad range of resources,
  focusing on the ones that are publicly available.
  In addition to providing information about the available linguistic resources,
  we present a set of recommendations,
  and identify gaps in the data available
  for conducting research and building applications
  in Turkish Linguistics and Natural Language Processing.
\end{abstract}

\section{Introduction}\label{sec:intro}

As in many other fields of science and engineering,
the data-driven methods have been the dominant approach
to natural language processing (NLP)
and computational linguistics (CL)
for the last few decades.
The recent (re)pop\-u\-lar\-iza\-tion of deep learning methods
increased the importance and need for the data even further.
Similarly,
the other subfields of theoretical and applied linguistics
have also seen a shift towards more data-driven methods.
As a result,
availability of large and high-quality language data is essential
for both linguistic research and practical NLP applications.
In this paper,
we present a comprehensive and critical survey of linguistic resources
for Turkish.

Turkish is a language spoken by over \num{80} million people,
mainly in Turkey, also having a significant number of speakers
in Cyprus, Europe, and Central Asia \citep{ethnologue}.%
\footnote{Throughout this paper, we use Turkish only for referring
  to the language variety spoken in modern Turkey
  and use of this variety in other countries/regions.
  Hence, this count does not include other Turkic languages,
  including ones mutually intelligible with Turkish.}
It exhibits a number of interesting linguistic characteristics
that are often challenging to handle in NLP applications
in comparison to the well-studied languages.

As a result,
the linguistic resources for Turkish are important
for building practical NLP applications for a large speaker community
as well as
for quantitative and computational approaches to linguistics,
including multilingual and cross-linguistic research.
Furthermore,
since Turkish is one of the largest and most well-studied languages
in the Turkic language family,
the resources we review below are potentially useful
for language transfer in NLP applications,
and as examples for resource and tool creation efforts
for the other Turkic languages.

Our survey mainly focuses on currently available resources
\citep[see][for a more historical account of Turkish corpora]{aksan2018}.
We also introduce a companion webpage which we update as new linguistic resources become available.%
\footnote{The web page is publicly available at \url{https://turkishnlp.github.io}.
  The current list was compiled mostly by our own efforts.
  However, we also welcome suggestions through
  a simple web-based form,
  and also through the GitHub repository associated with this URL.}
Our survey provides an overview of the available resources, giving details for the major ones, and aims to identify the areas where more effort is needed.
To our knowledge, this is the first survey of its kind on Turkish resources.
The most similar work is an edited volume of papers
on various NLP tasks for Turkish \citep{oflazer2018book}.
Unlike our work, however,
the focus is not the linguistic resources but NLP techniques and tools,
and most of the contributions are updated descriptions of the research published earlier.
A similar initiative to our companion website is
the recently announced Turkish Data Depository (TDD) project
\citep{safaya2022},\footnote{\url{https://tdd.ai/}.}
which aims to build a repository of data and models for Turkish NLP.
Our aim is collecting a more comprehensive list of pointers
which can be useful for both NLP and linguistic research,
while the TDD intends to store the actual data and the models
for NLP with a more practical purpose.

Our focus in this survey is linguistic data,
in particular, corpora and lexical resources.
We do not aim to describe the research questions,
methods and/or the results of these studies
but focus on describing the resources in detail.
We include resources that are potentially useful for NLP applications,
as well as for linguistic research.
We also do not focus on NLP tools explicitly,
such as data-driven part-of-speech (POS) taggers or parsers
and higher level tools or services that target non-technical audience
such as the web-based NLP pipelines \citep[e.g.,][]{eryigit2014,coltekin2015}.

The main contribution of the current paper is
a broad, comprehensive overview of the linguistic data available for Turkish
to enable linguists and NLP researchers/practitioners
to locate these resources easily.
We also identify missing or incomplete resources,
suggesting potential areas for future resource creation efforts.
We do not only offer a static survey,
but we intend to maintain a `living list' of resources and a repository
of publicly available linguistic data.

\section{Corpora}\label{sec:corpora}

This section surveys corpora available for Turkish.
We start with general-purpose, linguistically motivated corpora,
followed by corpora used for more specific purposes.

\subsection{Balanced corpora}\label{sec:balanced}

Since corpora collected from a single source (genre, domain)
contain many idiosyncratic aspects of its source,
the creation of balanced or representative corpora has been a major activity
in computational/corpus linguistics since
the earliest examples of linguistic corpora \citep[e.g.,][]{francis1979}.
There are two well-known balanced corpora for Turkish,
the Middle East Technical University (METU) corpus \citep{say2002}
and Turkish National Corpus \citep[TNC,][]{aksan2012}.

The METU corpus is the first balanced corpus released for Turkish.
The corpus consists only of written modality
sampled from \num{14} different text types including
novels, essays, research articles, travel articles, interviews, news,
newspaper columns, biographies and memoirs.
The corpus contains approximately \num{1000}
documents and \num{1.7}M tokens.%
\footnote{The numbers are based on a version obtained in 2015,
  which includes minor fixes to the first original release.}
The original release does not contain any linguistic annotations.
However, a number of annotation projects were
carried out on parts of this corpus
\citep[e.g.,][both discussed in Section \ref{sec:treebanks}]{oflazer2003,zeyrek2013}.
It is available free-of-charge for research purposes
after signing a license agreement.

The second balanced corpus is
the Turkish National Corpus \citep[TNC,][]{aksan2012}.
The TNC follows the design principles of the British National Corpus
\citep[BNC,][]{bnc}.
The corpus consists of \num{50}M words from texts collected from
books, periodicals, and various published and unpublished material.
It also includes a small `spoken text' portion
that consists of political speeches and news broadcasts.
The TNC contains texts from nine different domains (e.g.
fiction, scientific articles, art, opinions and editorials) and includes morphological annotations.
The corpus is not available for download
but it is accessible through a web interface.%
\footnote{Further information and the query interface is available
  from \url{https://www.tnc.org.tr/}.
}
A small part of the TNC is also used in constructing the BOUN Treebank
\citep[described below]{turk2022}.

\subsection{Treebanks and corpora with morphosyntactic annotation}
\label{sec:treebanks}

This section reviews primarily manually-annotated Turkish corpora
with ge\-ne\-ral-purpose \emph{linguistic} annotations,
as opposed to corpora annotated for a particular NLP task.
The majority of the corpora discussed below are treebanks,
however we also include a few other corpora
with morphosyntactic annotations.

Treebanks are important resources for linguistic research and applications.
Although they have been primarily used for training parsers in CL,
multiple levels of linguistic annotations available in treebanks
have also been beneficial for other NLP applications and linguistic research.
There has been a surge of interest in creating new treebanks
for Turkish in recent years.
Table~\ref{tbl:treebanks} presents the currently-available treebanks,
along with basic statistics.%
\footnote{%
    We only include manually annotated treebanks.
    All treebanks listed in the table are directly available for download,
    with the exception of ITU Web treebank,
    which requires a signed license agreement to be sent to the maintainers.
    All UD treebanks can be obtained through the project's webpage
    at \url{https://universaldependencies.org/}.
    Automatic conversion efforts or parsed corpora are not listed in the table.
}
Below, we provide a brief historical account of treebanks for Turkish.

\begin{table}
  \caption{A summary of currently available Turkish treebanks.
    The numbers in the table are based on our own counts
    on the most recent versions of the datasets.
    Not all information is reported in the respective papers,
    and there may be mismatches between the numbers reported
    in the papers and the released datasets.
  }\label{tbl:treebanks}
  \centering
  \begin{tabular}{lllS[table-format=5.0,table-number-alignment=right]S[table-format=6.0,table-number-alignment=right]}
    \toprule
    Treebank     && Type & {Sentences} & {Tokens} \\
    \midrule
    \weblink{https://ii.metu.edu.tr/metu-corpora-research-group}{METU-Sabanc{\i}} &\citet{oflazer2003}    & dep  & 5635  & 56396 \\
    \weblink{http://tools.nlp.itu.edu.tr/Datasets}{ITU Web}      &\citet{pamay2015}      & dep  & 5009  & 43191 \\
    \weblink{https://github.com/UniversalDependencies/UD_Turkish-GB}{UD-GB}        &\citet{coltekin2015tlt}& dep  & 2880  & 16803 \\
    \weblink{https://github.com/UniversalDependencies/UD_Turkish-PUD}{UD-PUD}       &\citet{zeman2017}      & dep  & 1000  & 16536 \\
    \weblink{https://github.com/UniversalDependencies/UD_Turkish-BOUN}{UD-BOUN}      &\citet{turk2022}       & dep  & 9761  &121214 \\
    \weblink{https://github.com/google-research-datasets/turkish-treebanks/}{TWT}          &\citet{kayadelen2020}  & dep  & 4851  & 66466 \\
    \weblink{https://github.com/olcaytaner/TurkishAnnotatedTreeBank-15}{Turkish-Penn-CS}
                 &\citet{yildiz2014}     & con & 9560  & 81419 \\
    \weblink{https://github.com/UniversalDependencies/UD_Turkish-Penn}{UD-Turkish-Penn}         &                    & dep  &  9560 & 87367 \\
    \weblink{https://github.com/UniversalDependencies/UD_Turkish-Tourism}{UD-Tourism}      &                    & dep  & 19750 & 92200 \\
    \weblink{https://github.com/UniversalDependencies/UD_Turkish-Kenet}{UD-Kenet}        &                    & dep  & 18700 &178700 \\
    \weblink{https://github.com/UniversalDependencies/UD_Turkish-FrameNet/}{UD-FrameNet}     &                    & dep  &  2700 & 19221 \\
    \bottomrule
  \end{tabular}
\end{table}

The first Turkish treebank is the METU-Sabanc{\i} treebank
\citep{atalay2003,oflazer2003}.  The METU-Sabanc{\i} treebank is a dependency
treebank including a selection of sentences from the METU corpus discussed
in Section \ref{sec:balanced}, and includes different text types of the original resource.  As an early
effort with relatively low funding, the treebank had various issues with
formatting and data quality \citep{say2011}.  Despite these issues, the
METU-Sabanc{\i} treebank was the only Turkish treebank over a decade.  There has
been a large number of reports of fixes over the years, but most fixes remained
unpublished, or even introduced other errors or unclear modifications to the
annotation scheme.  The most up-to-date version of this treebank is made
available through Universal Dependencies
\citep[UD,][]{nivre2016,demarneffe2021}
repositories based on a semi-automatic conversion
\citep{sulubacak2016} of a
version from Istanbul Technical University (ITU) and hence, named UD-IMST
(ITU-METU-Sabanc{\i} Treebank).
Even the latest version is reported to have a
large number of errors, carried over from earlier versions or introduced
along the way by many automated conversion processes \citep[see,
e.g.,][]{turk2019}.  \citet{burga2017} present a conversion of the same
treebank into another related framework, namely Surface-Syntactic Universal
Dependencies \citep[SUD,][]{gerdes2018}.
The paper states the intention to publish the resulting treebank,
but it is not available at time of this writing.

After a long time gap,
a growing number of new dependency treebanks have recently been released.
One of the new treebanks, ITU-Web treebank \citep{pamay2015},
contains user-generated text from the web.
It was annotated following the METU-Sabanc{\i} treebank annotation scheme,
and later converted to the UD  annotation scheme automatically.
The first treebank annotated directly using the UD framework is
by \citet{coltekin2015tlt}.
This treebank contains linguistic examples from a grammar book to increase
the coverage of different morphosyntactic constructions
while minimizing the annotation effort.
Two relatively larger and more recent dependency treebanks are
the Bo\u{g}azi\c{c}i University (BOUN) treebank \citep{turk2022}
and the \weblink{https://github.com/google-research-datasets/turkish-treebanks/}{Turkish web treebank} \citep[TWT,][]{kayadelen2020}.
The BOUN treebank annotates a selection of sentences from
the TNC \citep[see Section~\ref{sec:balanced}]{aksan2012} covering
a number of different text types.
The BOUN treebank is directly annotated according to the UD annotation scheme.
The TWT includes sentences from the web and Wikipedia.
The annotations in TwT deviate from the UD and the majority of the existing Turkish dependency treebanks.

Besides the monolingual treebanks above,
there have also been a few parallel treebanking efforts.
\citet{megyesi2008,megyesi2010} report automatically annotated
parallel dependency treebanks of Turkish, Swedish and English,
containing texts published in the forms of popular literature books. However, they have not been released publicly.
Another early attempt of parallel treebanking is
the constituency treebank described by \citet{yildiz2014,kara2020}.
This treebank includes translations of short sentences
(less than \num{15} words) from Penn Treebank \citep{marcus1993}.
The UD-PUD \citep{zeman2017} is part of a parallel dependency treebank effort
including \num{20} languages so far,
built on sentences translated predominantly from English.
The dependency annotations were performed by Google
with their own annotation scheme and automatically translated to
UD for the CoNLL multilingual parsing shared task
\citep{zeman2017}.
A different type of multilingual treebanking effort is
the UD-SAGT treebank,
which annotates \num{2184} spoken language utterances containing
Turkish--German code-switching treebank \citep{cetinoglu2019,cetinoglu2022}.
The treebank follows the UD framework.
Section~\ref{ssec:code-switching} provides further details about
the underlying dataset.

Version 2.8 of the UD treebanks, released in May 2021,
introduced four new Turkish treebanks from the same group.
One of these treebanks is the dependency version of
the Penn treebank translations \citep{yildiz2014}.
Others include a domain-specific tourism treebank,
and two treebanks annotating example sentences from two lexical resources
discussed in Section~\ref{sec:lexical} below.
The descriptions of the treebanks in the UD repositories indicate that
all four treebanks are manually annotated.
However, no formal descriptions of these treebanks
have been published at the time of writing.

As described above,
Turkish is relatively rich with respect to the quantity of available treebanks.
However, the need for improvement in terms of
the quality of annotations,
establishing standards
and resolving inconsistencies within and across treebanks
has been emphasized by multiple researchers
\citep[see, for example][for earlier discussions]{say2011,coltekin2016turcling,turk2022}.

An unusual, yet potentially useful
\weblink{https://odin.linguistlist.org/}{freely-available dataset}
with morphosyntactic annotation is ODIN \citep{lewis2006},
a multilingual collection of
examples from linguistics literature with interlinear glosses.
Although ODIN does not include full or uniform morphosyntactic annotations,
the glossed example sentences can be useful for linguistic research;
they may serve as test instances with interesting
or difficult linguistic constructions;
and they can be converted to a treebank
with less effort than that is required for annotating unanalyzed text.

There are also a few corpora that include only morphological annotations.
The most popular corpus with morphological annotations is
a \num{1}M token corpus disambiguated semi-automatically.
The exact procedure used for the disambiguation is unclear.
The corpus was introduced by \citet{hakkani-tur2002},
and made publicly available by later studies
on morphological disambiguation \citep{yuret2006,sak2011,dayanik2018}.
Another fully manually disambiguated dataset
consisting of \num{25098} words is reported in \citet{kutlu2013},
which can be obtained from the authors via email.

\subsection{Large-scale (unannotated) linguistic data collections}

Although well-balanced, representative corpora have been
at the focus of building corpora in corpus linguistics,
opportunistic large collections of linguistic data have also been useful
in CL/NLP tasks that require large datasets.
Furthermore,
the size and distribution restrictions on balanced corpora
often limits their use both for NLP applications,
and research on some linguistic questions
(e.g., if the questions are concerned with rare linguistic phenomena).
In this section,
we review some of the unannotated or automatically annotated corpora
that are either used in earlier literature,
or publicly accessible without major limitations.

The largest Turkish corpora available
are two large multilingual web-crawled datasets:
supplementary data released as part of
CoNLL-2017 UD parsing shared task \citep{zeman2017,ginter2017},
and the OSCAR corpus \citep{ortiz-suarez2019,ortiz-suarez2020}.
Both corpora are sentence shuffled to comply with the copyright laws.
The Turkish part of the CoNLL-2017 dataset contains
approximately \num{3.5} billion words.
The data is deduplicated,
and automatically annotated for morphology and dependency relations.
The data can be downloaded directly from the
\weblink{http://hdl.handle.net/11234/1-1989}{LINDAT/CLARIN repository}.
The OSCAR corpus is available as raw, and deduplicated versions.
The Turkish section contains over \num{3} billion words after deduplication.
\weblink{https://oscar-corpus.com/}{The OSCAR corpus} can be obtained
after creating an account automatically.
The publicly available data does not include any meta information,
and the order of the sentences is destroyed by shuffling.
However, the webpage of the OSCAR corpus includes a form to request
original data without sentence shuffling.

Another popular, relatively  large Turkish corpus is the BOUN corpus
\citep{sak2008}.
The corpus contains approximately \num{500}M tokens
collected from two major online newspapers and other webpages.
Although it is used in many studies, it is not clear how to access this corpus.

A relatively large, and easily accessible data source is
the multilingual
\weblink{https://corpora.uni-leipzig.de/}{Leipzig Corpora Collection}
\citep{quasthoff2014}.
The Turkish section contains over \num{7}M sentences
(approximately \num{100}M words) of news, Wikipedia and web crawl.
The Leipzig corpora are also sentence shuffled.
Web-crawled data also contains smaller parts crawled
from Turkish-language web sites published in Cyprus and Bulgaria.

The Turkish parliamentary corpus released
as part of the \weblink{https://www.clarin.eu/parlamint}{ParlaMint project}
\citep{erjavec2021,erjavec2022}
contains the transcripts of the Turkish parliament (2011--2021),
including approximately \num{43}M words
from \num{303505} speeches delivered at the main proceedings of the parliament.
The data also contains speaker information
(name, gender, party affiliation)
and automatic annotations including morphology,
dependency parsing and named entities.

Another relatively large (approximately \num{10}M words),
freely accessible corpus is
\weblink{https://www.kaggle.com/alvations/old-newspapers}{the Kaggle old news dataset}.%
\footnote{The corpus is not described in any earlier publication.
  Throughout this survey,
  we cite the papers describing each resource, if one is available,
  otherwise provide a hyperlink to the resource.
  Links to all available resources are
  provided in the companion webpage at \url{https://turkishnlp.github.io}.
}
This is a multilingual collection from well-known news sites.
The data also includes publication date of the article and the source URL
of the document.

\weblink{https://tscorpus.com/corpora/}{The TS Corpus}
\citep{sezer2013,sezer2017}
is also a large collection of corpora with a web interface.
The collection contains some corpora released earlier
(e.g., the BOUN corpus discussed in Section \ref{sec:balanced})
as well as sub-corpora collected by the authors.
The authors report over \num{1.3} billion tokens in \num{10} sub-corpora
from various text sources and various levels of (automatic) annotation.
The corpus is served via a web-based query interface,
and, to our knowledge, the full corpus is not publicly available for download.

Another relatively small, but potentially interesting unannotated dataset
is a compilation of \num{6844}
\weblink{https://stars.bilkent.edu.tr/turkce/}{%
essays on creative writing classes} by Turkish university students
between 2014--2018.
The essays (approximately \num{400}K words) are published
on the course webpage as PDF files.

\subsection{Corpora with discourse annotation}
There are two corpora that are annotated for discourse markers in Turkish.
The first one, Turkish Discourse Bank \citep[TDB,][]{zeyrek2013},
includes roughly \num{400}K words across various written genres
in the METU corpus (Section~\ref{sec:balanced}).
The corpus is annotated based on explicit connectives and their two arguments.
The TDB is available for academic use through email.
\citet{zeyrek2018,zeyrek2020}, on the other hand,
focus on annotating discourse markers in the transcripts of TED talks
in six languages (i.e., English, German, Polish, European Portuguese, Russian and Turkish).
The Turkish corpus measures \num{5164} words. The annotation tasks in each language were carried out according to the Penn Discourse Treebank (PTDB) guidelines.
The corpus was annotated for five discourse relation types (i.e., explicit connectives, alternative lexicalizations, implicit connectives, no relation) and five top-level senses (i.e., temporal, comparison, expansion, contingency, hypophora).
The annotated corpus is \weblink{https://github.com/MurathanKurfali/Ted-MDB-Annotations}{freely available}.

\subsection{Word sense disambiguation corpora}

\begin{table}
  \caption{A summary of WSD resources. The `Additional'
  column mentions additional annotations, namely, morph: POS tags and morphology, dep: dependency, con: constituency.
  }\label{tbl:wsd}
  \centering
  \begin{tabular}{*{4}{l@{\hspace{6pt}}}S[table-format=4.0]S[table-format=5.0]}
    \toprule
    Resource & & Type & Additional & {Samples} & {Sent.} \\
    \midrule
    METU & \citet{orhan2007} & lexical sample &  morph, dep & 26 & 5385  \\
	ITU & \citet{ilgen2012} & lexical sample &  - & 35 & 3616  \\
	I\c{s}{\i}k & \citet{akcakaya2018}  & all-words  & morph, con & 7595 & 83474 \\
    \bottomrule
  \end{tabular}
\end{table}

The word sense disambiguation (WSD) task has been defined in two ways: lexical sample and all-words. The lexical sample task aims to disambiguate a restricted set of ambiguous words in their context. The all-words variant, on the other hand, disambiguates all words of a given input. Turkish has resources for both variants.

The first WSD dataset for Turkish is created as part of a SemEval 2007 task and opts for the lexical sample variant \citep{orhan2007}. \num{26} unique lexical samples are tagged for their senses, and each sample is tagged in about \num{100} sentences. The corpus used for the annotation is the METU-Sabanc{\i} Treebank, hence the WSD dataset is already accompanied with morphosyntactic annotations. The WSD annotation adds fine-grained senses from the dictionary of Turkish Language Association (TDK), coarse-grain senses, which are a set of semantically closest fine-grained senses, and three levels of ontology.
The website link provided in the paper for obtaining the resource is not accessible.
%The resource is not available
%since the university hosted it is not active anymore. 

\citet{ilgen2012} also employ the lexical samples approach but choose their words among the most ambiguous words based on a frequency list \citep{goz2003}.
There are \num{35} lexical samples in total and each sample is annotated in at least \num{100} sentences. The corpus was collected from well-known websites on news, health, sports, and education in Turkish. The word senses come from the TDK dictionary (though the authors eliminated some senses that are infrequent in online resources).
The availability of the resource is unclear.

The first all-words WSD resource for Turkish annotates a set of sentences that contains translations of Penn Treebank sentences up to \num{15} tokens
(the treebank is described in Section~\ref{sec:treebanks}).
\citet{akcakaya2018} annotates the dependency version of the treebank
as an all-words WSD resource.
Therefore, the sentences also include morphosyntactic annotations. As in other resources, the sense information comes from the TDK dictionary.\footnote{Note that it is the same dictionary, yet different versions.} In total, there are \num{7595} unique lexical samples to disambiguate in a corpus of \num{83473} tokens. \SI{77}{\percent} of these unique samples
are nouns, followed by verbs and adjectives.
The website link provided in the paper for obtaining the resource is not accessible.
The statistics for WSD resources are given in Table \ref{tbl:wsd}.

\subsection{Corpora of parent-child interactions}

Language acquisition has been a major interest in modern lin\-guis\-tics,
where Turkish also received a fair amount of attention
because of a rather interesting learning course observed by young learners,
for example, an early and error-free acquisition of case morphology \citep{xanthos2011}.
\weblink{https://childes.talkbank.org/}{The CHILDES database}
\citep{macwhinney1985} contains two freely-available Turkish datasets
with transcriptions of parent--caregiver interactions.
The first dataset \citep{aksu-koc1985} contains transcripts of
\num{54} sessions consisting of interactions with \num{33} children between
\num{28} to \num{56} months of age.
The second dataset \citep{turkay2005,altinkamis2012}
contains transcriptions of \num{15} recordings with the same child
between ages \num{16} months to \num{28} months.
Both corpora mark speakers,
and include some extra-linguistic information.
The latter corpus also includes morphological annotation of
a subset of the child utterances.
A larger and more recent child-language dataset is reported in \citet{moran2015}.
However, the Turkish section of this corpus was not released as of this writing. \citet{rothweiler2011} has also released a `Turkish-German successive bilinguals corpus' which contains \num{94} longitudinal spontaneous speech samples by Turkish-German bilingual children (7--28 months-old) recorded between 2003--2008. Part of the data could be viewed for research purposes after obtaining a password.

\subsection{Social media text normalization corpora}

Normalization of social media text is an important first step
in many NLP applications,
where ill-formed words or phrases are replaced (or associated)
with their normal forms.
The definition of `ill-formed' text is debatable
and text normalization in social media hinders
analyzing social aspects of language use from
a computational sociolinguistic point of view \citep{Nguyen2016,eisenstein2013}.
However, normalization datasets enable the use of tools created
for formal/standard language,
and non-destructive text normalization is also helpful in analyzing
interesting aspects of non-standard language use by individuals or groups.
We review corpora for normalization purposes here,
for lexical resources for the same purpose, see Section~\ref{ssec:app-specific-lexicon}.

\citet{eryigit2017} report a `big Twitter dataset' (BTS)
for normalization which consists of
\num{26149} tweets, as well as using IWT
(see Section~\ref{sec:treebanks}) as a source of normalization data.
The BTS contains \num{57088} manually normalized tokens
out of a total of \num{385568}.
In IWT, \num{5101} tokens (out of \num{39152} are normalized.
The datasets are available from
\weblink{http://tools.nlp.itu.edu.tr/Datasets}{the group's webpage}
after signing a license agreement.
\citet{colakoglu-etal-2019-normalizing} introduced
another normalization test set of \num{713} tweets
(\num{7948} tokens, \num{2856} normalized).
The dataset is available via \weblink{https://bitbucket.org/robvanderg/multilexnorm/src/master/data/tr/}{W-NUT 2021 Shared Task on Multilingual Text Normalization}.
A more recent Twitter normalization data
consisting of \num{2000} sentences was introduced in \citet{koksal2020}.
\num{6488} out-of-vocabulary (OoV) tokens (out of \num{16878}) identified using
lexical resources were manually annotated
(below \SI{10}{\percent} of the OoV tokens are well-formed,
e.g., foreign names or neologisms).
The dataset is available through \weblink{https://github.com/atubakoksal/annotated_tweets}{a GitHub repository}.
Besides these monolingual resources,
\weblink{https://github.com/ozlemcek/TrDeNormData}{a normalization dataset for Turkish--German} is also available \citep{van-der-goot-etal-2021-lexical}.
This dataset is a revised version of the data from \citet{cetinoglu2016}
for normalization by employing token-level alignment layers and adapting existing language ID and POS tags for these new layers.

\subsection{Corpora for named entity recognition}

Named entity recognition (NER) for Turkish has been studied
by diverse groups of researchers with a few publicly available datasets.
\citet{tur2003} is one of the first to study NER in Turkish
with a dataset compiled from newspaper articles
over approximately one year (1997--1998).
The dataset is annotated for ENAMEX (person, location, organization)
named entity types.
The dataset has been the standard benchmark for many subsequent studies,
with some changes along the way.
Original article reports a dataset of approximately \num{1}M words.
The version of the dataset as used by \citet{yeniterzi2011}
consists of approximately \num{500}K words
with \num{37189} named entities (\num{16291} person, \num{11715} location \num{9183} organization).
This version of the data can be obtained through email.
\citet{celikkaya2013} report three additional datasets
covering different text sources, namely,
a computer hardware forum,
orders to a speech assistant,
and Twitter.
The data is also annotated for NUMEX entities (numerical expressions).
\citet{seker2017} report an annotation effort partially
based on the datasets reported in \citet{celikkaya2013} and \citet{tur2003},
but also annotating the IWT (described in Section~\ref{sec:treebanks}).
The datasets are available from
\weblink{http://tools.nlp.itu.edu.tr/Datasets}{the group's webpage}
after signing a license agreement.
\citet{eken2015} also report an additional \num{9358} tweets annotated
similar to \citet{celikkaya2013}.
However, availability of this dataset is unclear.

\citet{kucuk2014} and \citet{kucuk2019} report
two Twitter datasets of \num{2320} and \num{1065} tweets, respectively.
These datasets are annotated for
person, location, organization, date, time, money and misc
(e.g., names of TV programs, music bands),
and publicly available through
\weblink{https://github.com/dkucuk/}{the authors' GitHub repositories}.
Another, more recent, 
\weblink{https://github.com/SU-NLP/SUNLP-Twitter-NER-Dataset}{NER data set}
annotating \num{5000} tweets was released by \citet{carik2022}.

\subsection{Code-switching corpora}\label{ssec:code-switching}

Code-switching refers to mixing
more than one language in written and spoken communication and it is quite common in multilingual settings (e.g., immigration contexts, India, Africa etc.).

\citet{nguyen2013} and \citet{papalexakis2014} report analyzing code-switching (e.g., Turkish-Dutch) in online fora for automatic language identification and a prediction task but this data set is not publicly available. 

\citet{cetinoglu2016lrec} released a
\weblink{https://github.com/ozlemcek/TrDeNormData}{Turkish--German Twitter corpus} which is annotated with language IDs. The dataset consists of \num{1029} tweets that are automatically collected, semi-automatically filtered, and manually annotated. Each tweet contains at least one code-switching point, the tweets are normalized and tokenized before adding language IDs.
\citet{cetinoglu2016} added POS tag annotations
to the same dataset following UD guidelines.
A spoken corpus of interviews with Turkish--German bilinguals
was presented by \citet{cetinoglu2019,cetinoglu2022}.
The audio files are annotated with sentence and code-switching boundaries. Sentences that contain at least one code-switching point are transcribed and normalized to their orthographic representation. The resulting \num{2184} sentences are annotated with language IDs following \citep{cetinoglu2017}, and with lemmas, POS tags, morphological features, and dependency relations following the UD framework. \weblink{https://github.com/UniversalDependencies/UD_Turkish_German-SAGT}{The treebank version of the dataset} is available in the Universal Dependencies repositories, the audio files and aligned transcriptions are available to researchers after signing a license agreement.
\citet{yirmibesoglu2018} worked on detecting code-switching in Turkish--English social media posts. The data is claimed to be available but it was not found on the website link suggested in the paper.

\weblink{https://www.uni-potsdam.de/de/daf/projekte/multilit}{The MULTILIT project} \citep{schroeder2015} focuses on multilingual children and adolescents of Turkish and Kurdish background living in Germany and France. The corpora they collected include Turkish oral monologues (and their transcription), and written text produced by bilingual students. A subset of the corpus is annotated with POS tags, morphological features and partial syntactic structures, as well as markers showing deviations from standard language use. The data is not publicly available.
\weblink{https://www.linguistik.hu-berlin.de/en/institut-en/professuren-en/rueg}{The RUEG project} aims at similar goals at a larger range of age groups, and investigates bilingual speakers of Russian, Turkish and Greek background in Germany and the U.S., bilingual speakers of German in the U.S., as well as monolingual speakers of these languages in respective countries. As part of their collection there are Turkish corpora collected in Germany (\num{1197} sentences) and in Turkey (1418 sentences), publicly available as audio files and annotated transcriptions \citep{wiese2020}. The lemmas, POS tags, and morphological features are manually annotated, dependencies are automatically predicted. All layers follow the UD framework except the fine-grained POS tags which follow the MULTILIT project.

\subsection{Parallel corpora}\label{sec:parallel}

\begin{table}
  \caption{A selection of parallel corpora available for Turkish.
    The third column lists the languages in each corpus
    (numbers include Turkish),
    for massively parallel corpora Turkish may not be aligned
    to all languages.
    The number of sentences indicates the number of Turkish sentences
    in the particular corpus.
    The number of actual aligned sentences vary depending on the target language.
    All numbers are based on the corpora as available
    from OPUS parallel corpora collection \url{http://opus.nlpl.eu/}.
  }\label{tbl:parallel-corpora}
  \begin{tabular}{*{3}{l@{\hspace{6pt}}}S[table-format=9.0,table-number-alignment=right]}
  \toprule
  Corpus        & Text type  & Languages & {Sentences} \\
  \midrule
  Bianet \citep{ataman2018}  &News& English, Kurdish &   61472\\
  Bible                      &Religious     & Multiple (\num{102})  &48500\\
  EU book shop               &EU texts      & Multiple (\num{48})   &33398\\
  GlobalVoices               &News          & Multiple (\num{92})   &8796\\
  JW300 \citep{agic2019}     &Religious     & Multiple (\num{380})  &535353\\
  OpenSubtitles              &Subtitles     & Multiple (\num{62})  &173215360\\
  QED \citep{abdelali2014}   &Educational   & Multiple (\num{225})  &753343\\
  SETimes \citep{tyers2010}  &News          & Balkan (\num{10})       &1776431\\
  TED talks                  &Subtitles     & English               &746857\\
  Tanzil                     &Religious     & Multiple (\num{42})   &105597\\
  Tatoeba                    &Misc          & Multiple (\num{359})  &746857\\
  Wikipedai \citep{wolk2014} &Wikipedia     & English, Polish       &175972\\
  infopakki                  &Informational & Multiple (\num{12})   &50909\\
  \bottomrule
  \end{tabular}
\end{table}

Parallel, aligned corpora in multiple languages are
essential for machine translation (MT) as well as
multilingual or cross-lingual research.
A number of parallel corpora including Turkish have been reported
in some of the earlier works on MT between Turkish and mainly English
\citep[e.g.,][]{durgar-el-kahlout2010,oflazer2018mt,durgar-el-kahlout2019}.
Similarly,
shared tasks which included Turkish as one of the languages,
such as two IWLST shared tasks \citep{paul2010,cettolo2013},
and WMT shared tasks between 2016 and 2018 \citep{bojar2016},
also provided data for use during the shared tasks. However, none of these resources are available,
nor are there clear procedures to obtain these datasets.
In this review we only list
the resources available (for at least for non-commercial, research purposes)
in detail.

Almost all publicly available parallel corpora that include Turkish
are available from \weblink{http://opus.nlpl.eu/}{the OPUS corpora collection}
\citep{tiedemann2012opus}.
A selection of publicly available corpora are listed in
Table~\ref{tbl:parallel-corpora}
(except the parallel treebanks discussed in Section~\ref{sec:treebanks}).
The table does not list corpora of public software localization texts
and some of the other small corpora available through OPUS.
The sizes, text types and the target languages vary considerably.
This list of resources, to our knowledge,
are not used widely by researchers interested in
machine translation to/from Turkish.

Another active area of machine translation
is translation between Turkic languages
(e.g., \cite{hamzaoglu1993,altintas2001,tantuug2007,gilmullin2008,gokirmak2019};
see \cite{tantug2018} for a recent summary).
Similar to the Turkish--English translation studies,
the resources specifically built for the purpose are scarce,
and even if they are reported in the literature,
to our knowledge, no specific corpora build for translation
between Turkic languages were released.%
\footnote{Except \citet{gokirmak2019},
  who state the intention to release their data pending copyright clearance,
  most papers do not include intentions of sharing their data.
  }
Except for small samples in
\weblink{https://apertium.org/}{Apertium} repositories \citep{forcada2011},
the corpora build with large-scale
parallel text collections
(e.g., ones listed in Table~\ref{tbl:parallel-corpora})
seem to be the only easily obtainable resource for
studies requiring parallel corpora between Turkic languages.

\subsection{Corpora for sentiment and emotion}

\citet{demirtas2013} introduced two Turkish datasets consisting of
movie and product reviews.
The movie reviews, scraped from a popular Turkish movie review site,
contain \num{5331} positive and \num{5331} negative sentences.
The product reviews data, scraped from an online retailer web site,
consists of \num{700} positive and \num{700} negative reviews.
The labels are assigned based on the scores assigned
to the movie or the product by the reviewer.
The datasets are available at
\weblink{https://www.win.tue.nl/~mpechen/projects/smm/\#Datasets}{%
  the author's web site}.

\citet{kaya2013} used a balanced corpus of \num{400} newspapers columns
from \num{51} journalists labeled for positive and negative sentiment.
The study also reports a Twitter corpus of \num{123074} tweets (not labeled).
\citet{turkmenoglu2014} also report multiple datasets, consisting of
\num{20244} movie reviews, \num{4324} tweets and \num{101346} news headlines.
The tweet dataset was annotated with three-way classes
(positive, negative, neutral).
Similar to other studies, movie reviews are labeled them based on the scores
assigned by the reviewers. However, it is not clear how the authors labeled the headlines corpus and used it for the presented research. 
\citet{yildirim2014sa} report another manually annotated
Twitter dataset of \num{12790} tweets,
labeled as positive (\num{3541}) negative (\num{4249}) and neutral (\num{5000}).
None of these publications indicate the availability of the corpora introduced.
\citet{hayran2017} present another dataset of \num{3200} tweets.
The data is labeled (negative or positive)
based on the emoticons in the messages.
The dataset is available through email.

\citet{boynukalin2012} has investigated emotions in Turkish through two datasets.
The first dataset is a translation of a multilingual emotion corpus
\citep[ISEAR,][]{scherer1994}
into Turkish where the participants are asked to describe experiences
associated with a given set of emotions (e.g., joy, sadness, anger).
Although the original study describes seven emotions,
the authors focused on four of them in Turkish and they have identified \num{4265} short texts in total.
The second dataset consists of \num{25} fairy tales in Turkish collected across various websites on the web. The emotions in this dataset were labeled based on intensity (low, medium, high) at the sentence and paragraph levels.
\citet{demirci2014} analyzed the emotions in a dataset of \num{6000} tweets,
and labeled based on the hashtags they contain
as anger, fear, disgust, joy, sadness, surprise.
The availability of these two datasets is unclear.
A more recent emotion dataset, TREMO, based on the ISEAR corpus
is presented by \citet{tocoglu2018}.
Instead of translating the original texts,
\citet{tocoglu2018} follow the methodology used to collect the ISEAR corpus,
and collect \num{27350} entries from \num{4709} individuals
describing memories and experiences related to six emotion categories.
\citet{tocoglu2019} built a dataset consisting of \num{195445} tweets
automatically labeled with these emotion categories based on a lexicon
(see Section~\ref{ssec:app-specific-lexicon}) extracted from the TREMO dataset.
Both of these datasets are
\weblink{http://demir.cs.deu.edu.tr/downloads/}{available online}
for non-commercial use.

\subsection{Speech and multi-modal corpora}

As in other languages,
speech corpora or other forms of multi-modal datasets (e.g., video)
are scarce in comparison to text corpora.
The only linguistically motivated speech corpus creation effort
seems to be the \weblink{https://std.metu.edu.tr/}{Spoken Turkish Corpus}
\citep[STC,][]{ruhi2010,ruhi2012}.
Although an initial sample consisting of
\num{20} recordings, \num{4514} utterances and  \num{16107} words
was released in 2010,
the full corpus is still not available.

Easily-accessible Turkish speech corpora are generally
parts of multilingual corpus creation efforts.
Notable examples include
\weblink{https://commonvoice.mozilla.org/en/datasets}{Common Voice}
\citep{ardila2020},
and \weblink{https://github.com/NTRLab/MediaSpeech}{MediaSpeech}
\citep{kolobov2021}.
The Common Voice dataset is an ongoing data collection effort by
Mozilla Foundation.
The project collects audio recordings of a set of sentences and phrases
in multiple languages. 
The January 2022 release
includes over \num{68} hours of recordings from \num{1228} Turkish speakers.
The MediaSpeech dataset includes \num{10} hours of speech recordings
(\num{2513} short segments less than \num{15} seconds each)
with transcriptions from two news channels.
\weblink{https://ict.fbk.eu/must-c/}{MuST-C} \citep{di-gangi2019,cattoni2021}
is a multilingual corpus of TED talks including Turkish transcripts,
but the audio data is only in English.

The majority of the other speech datasets are collected/created
within practical speech recognition/processing projects
\citep[see][for a recent review of Turkish speech recognition]{arslan2020}.
The speech corpus introduced in \citet{mengusoglu2001}
consists of broadcast news
and a set of sentences from news read by multiple speakers.
Another early speech corpora collection is OrienTel-TR \citep{ciloglu2004},
Turkish part of the multilingual OrienTel project \citep{draxler2003},
collecting phone recordings of pronunciations
of a selected set of words and phrases.
\citet{arisoy2009bn} report a larger dataset of broadcast news,
and a dataset of \num{38000} hours of call center recordings
is reported by \citet{haznedaroglu2014}.
A recent speech corpus,
consisting of movies with aligned subtitles,
and read speech samples are reported by \citet{polat2020}.
The availability of corpora listed in this paragraph is unclear.

\citet{salor2007} report a spoken corpus of \num{2462} sentences,
read by \num{193} speakers with varied ages and backgrounds.
Another, similar but smaller set of recordings are available through
GlobalPhone corpus \citep{schultz2013},
which is a collection of parallel sentences from \num{20} languages including Turkish.
Another interesting dataset where native speakers were recorded
while reading parts of dialogues in the ATIS corpus
\citep{hemphill1990} is reported in \citet{upadhyay2018}.
These corpora are available for purchase through the LDC or the ELRA.

\citet{topkaya2012}
report a dataset of audio/video recordings
in which \num{141} Turkish speakers pronounce selected numbers, names,
phrases and sentences in a controlled environment.
Finally, it is also worth mentioning
the Turkish--German spoken code-switching treebank
described in Section~\ref{ssec:code-switching} contains
aligned audio recordings of Turkish--German bilinguals.
Both datasets can be obtained by contacting the authors.

\subsection{Corpora for question answering}

Although a highly applicable and popular area,
there have been relatively few Turkish resources available
for question answering (QA) until recently.
Early QA work on Turkish include short lists of question--answer pairs
without the context including the answer.
For example \citet{amasyali2005}
report the use of a \num{524} question--answer pairs.
However, to our knowledge none of these datasets are made available.
Similarly, \citet{pala2009} includes \num{105} factoid questions and their answers as part of her thesis manuscript.
\citet{longpre2020} present
\weblink{https://github.com/apple/ml-mkqa}{a freely-available dataset}
containing human translations of \num{10000} question--answer pairs
sampled from the Natural Questions dataset \citep{kwiatkowski2019}
to \num{25} languages including Turkish.
\weblink{https://github.com/deepmind/xquad}{Another multilingual QA set}
released by \citet{artetxe2020}
includes a \num{1190} human-translated question--answer pairs from
Stanford Question Answering Data Set \citep[SQuAD,][]{rajpurkar2016}.
In a more recent study, \citet{gemirter2020} report
both a new domain-specific dataset as well as
an automatic translation of SQuAD.
The availability of this dataset is unclear.

\subsection{Other corpora for specific applications}\label{sec:other-corpora}

The subsections above survey
the areas where a relatively large number of resources are available.
In this subsection, we review other areas where there are relatively
few resources,
either because it is a relatively new area,
or because there has not been enough interest
in the Turkish CL community.

\topic{%
  Offensive or aggressive language online} has been a concern
since the early days of the Internet \citep{lea1992}.
With the increasing popularity of social media,
and because of the regulations introduced
against certain forms of offensive language such as hate speech online,
there has been a surge of interest in automatic detection of
various types of offensive language.
Currently, there are two Turkish corpora related to offensive language.
The cyberbullying corpus by \citet{ozel2019} is a manually annotated
corpus of \num{15658} comments collected from multiple social media sites.
This dataset is not available.
The corpus reported in \citet{coltekin2020} is
a general offensive language corpus hierarchically annotated
according to OffensEval guidelines \citep{zampieri2019}.
This corpus is
\weblink{https://coltekin.github.io/offensive-turkish/}{publicly available}
and consists of \num{36232} manually annotated tweets.
In addition,
two recent hate speech date sets were released by research groups at
\weblink{https://github.com/avaapm/hatespeech}{Aselsan}
\citep{toraman2022},
at the
\weblink{https://github.com/verimsu/Turkish-HS-Dataset}{Sabancı University}
\citep{beyhan2022}.

\topic{Natural language inference} (NLI) attracted
considerable interest recently.
\weblink{https://github.com/facebookresearch/XNLI}{The cross-lingual NLI dataset}
\citep[XNLI,][]{conneau2018},
includes \num{7500} premise--hypothesis pairs created for English,
and translated to Turkish as well as \num{13} other languages.
More recently, \citet{budur2020} released
\weblink{https://github.com/boun-tabi/NLI-TR}{a dataset}
consisting of automatic translations of Stanford NLI \citep[SNLI,][]{bowman2015}
and MuliNLI \citep{williams2018} datasets,
consisting of approximately \num{570000} and \num{433000} sentence pairs,
respectively.
A small part of the SNLI data (\num{250} sentence pairs) was
also translated to Turkish earlier for a SemEval-2017 task \citep{cer2017}.
The data is available from the \weblink{http://ixa2.si.ehu.eus/stswiki/index.php/Main_Page}{SemEval-2017 multilingual textual similarity shared task website}.
All NLI datasets listed above are publicly available.

\topic{Summarization} datasets for Turkish are
also mostly from multilingual corpora collection efforts
\citep[e.g.,][]{ladhak2020,scialom2020}.
Almost all work on summarization of Turkish texts we are aware of
\citep[e.g.,][]{kutlu2010,ozsoy2011}
rely on automatic ways to obtain texts and their summaries.
However, the availability of these corpora is not clear.

\topic{Paraphrasing} corpora have interesting applications such as
machine translation and determining semantic similarity.
Two paraphrasing corpora in Turkish are introduced in
\citet{demir2012} and \citet{eyecioglu2016}.
The former study reports an unpublished (work-in-progress) corpus
of \num{1270} paraphrase pairs but it is not publicly available yet.
The latter study reports
\weblink{https://osf.io/wp83a/}{a publicly-available corpus}
of \num{1002} paraphrase pairs
which also includes human-rated semantic similarities of the sentence pairs.
Another textual similarity dataset
created by automatic translation of
the English STS benchmark \citep{cer2017} is  
\weblink{https://github.com/verimsu/STSb-TR}{published} by \citet{beken-fikri2021}.

\topic{Text categorization} or topic modeling studies in Turkish
often use opportunistic labeling of the topics published in newspaper sections (e.g., politics, economics, sports).
Although there are many studies reporting such datasets,
they are rarely made publicly available.
We only note one
\weblink{https://github.com/denopas/TTC-3600}{publicly available}
corpus by \citet{kilinc2017}
which has become a common benchmark data for later studies.
This corpus consists of \num{3600} news feeds (RSS) obtained from online newspapers in \num{6} categories.

Similar to text categorization, stylometry related studies also typically use
newspaper columns scraped from online newspapers,
and the corpora are not made available publicly
(possibly also due to copyright restrictions).
Exception we are aware of are
\weblink{http://www.kemik.yildiz.edu.tr/}{a few datasets available from}
Y{\i}ld{\i}z Technical University NLP group \citep{amasyali2006,turkouglu2007}
and \weblink{https://cloud.iyte.edu.tr/index.php/s/5DhqdlUCCdB60qG}{the publicly available dataset}
of Twitter gender identification corpus by \citet{sezerer2019},
which contains \num{5292} users with more than \num{100} tweets each
manually labeled for gender.

Coreference resolution is another task for which the quantity of resources
available is rather small.
Earlier work on coreference resolution \citep{kucuk2007,kucuk2008}
report the use of annotated corpora without indication of availability.
In the only \weblink{https://bitbucket.org/knowlp/marmara-turkish-coreference-corpus/}{publicly available} corpus with coreference annotation,
\citet{schuller2017} annotate all sentences of METU-Sabanc{\i} treebank
(described in Section~\ref{sec:treebanks}) for coreference.

We also note two large multilingual COVID19-related tweet collections
by \citet{qazi2020} and \citet{abdul-mageed2021}.
\weblink{https://crisisnlp.qcri.org/covid19}{The first corpus}
focuses on tweets geo-location in many languages.
Although the number of tweets in Turkish is not specified,
the total number of tweets is about half a billion.
\weblink{https://github.com/UBC-NLP/megacov}{The second corpus} includes
\num{28.5}M Turkish tweets with COVID-19 related keywords.
Both COVID-19 datasets are available as tweet IDs.
\citet{kartal2020} presents a dataset of Turkish \num{2287} tweets
labeled whether they are worth fact checking or not.
The dataset is available through
\weblink{https://github.com/YSKartal/TrClaim19}{a GitHub repository}.

Last but not the least, we note two sign-language corpora.
The first corpora of Turkish sign language was introduced by \citet{camgoz2016},
and contains sentences and phrases from finance and health domains.
\citet{eryigit2020} present a Turkish sign language
with morphological and dependency annotations,
as well as parallel sentences in Turkish.
The availability of these two corpora is unclear.
\citet{sincan2020} describe a publicly available sign language corpus.
However \weblink{https://cvml.ankara.edu.tr}{the link} provided 
in the article is not active at the time of this writing.

\section{Lexical Resources}\label{sec:lexical}

In this section we describe large lexicons and lexical networks
that are built either as standalone projects or as part of multilingual collections.
The majority of these lexicons also provide various levels of annotations and in multilingual
cases, they usually have a mapping to the other languages of the collection.

\subsection{Lexicons, word lists}
\label{ssec:lexicons}

\citet{inkelas2000} aim at creating a 
\weblink{http://linguistics.berkeley.edu/TELL/cgi-bin/TELLsearch.cgi}{Turkish Electronic Living Lexicon} (TELL)
that reflects actual speaker knowledge. The lexicon they built consists of \num{30000}
lexemes from dictionaries and place names.
Nouns are inflected for five forms and verbs are for three, more than half also
have morphological roots. All entries have phonemic transcriptions, \num{17500} of them
also have pronunciations.
Moreover, \num{11500} entries are annotated with their etymological source language.
It is possible to search the whole lexicon via a webpage
which also offers an email address to access the database.
LC-STAR \citep{fersoe2004} is a collection of lexicons for
speech translation between \num{13} languages including Turkish.
The Turkish lexicon consists of \num{59213} common words
(in sport, news, finance, culture, consumer information, and personal communication domains)
and \num{43500} proper names of persons, places, and organizations.
The data has been originally released via ELRA but currently it is not available in
their catalog. 

\weblink{https://babelnet.org/}{BabelNet} \citep{navigli2012} is a semantic network covering \num{284} languages, It is
created using WordNets,
Wikipedia, and machine translation.
The project's webpage
offers a search interface for end users
and APIs for programmers.
\weblink{https://vocab.panlex.org/tur-000}{PanLex} \citep{kamholz2014} builds translation lexicons for over \num{5700} languages
by utilizing their dictionaries and other multilingual resources such as WordNets.
The project's webpage lists collected lexicons
and available resources for each language.
However, most links for Turkish seem to be broken.
 While PanLex is the largest
among mentioned lexicons, it should be noted that some non-Turkish entries are marked as Turkish.
The lexicons, their number of lexemes, and additional annotations are summarized in Table~\ref{tbl:lexicons}.
\begin{table}
  \caption{The statistics for Turkish large-scale lexicons. The `Additional'
  column mentions additional annotations. `etymo.' stands for etymological source.
  }\label{tbl:lexicons}
  \begin{tabular}{llS[table-format=6.0,table-number-alignment=right]p{0.35\textwidth}}
    \toprule
    Lexicon     &  & {Lexemes} & Additional \\
    \midrule
    TELL &\citet{inkelas2000}      & 30000 & phonemic transcriptions, roots, inflected forms, etymo. \\
    LC-STAR & \citet{fersoe2004}   & 104513 &  phonetic transcriptions\\
    BabelNet      & \citet{navigli2012}    & ? &  translations, semantic relations\\
    Panlex      &  \citet{kamholz2014}   & 242635 & translations \\
    \bottomrule
  \end{tabular}
\end{table}

Inflectional and derivational lexicons focus on the morphosyntactic representations
of words.
The UniMorph project \citep{sylak-glassman2015,kirov2016} aims at building a universal
schema for morphological representation of inflected forms.
So far,  over \num{120} languages are annotated (based on their webpage)
with their features in a combination of automatic extractions from Wiktionary and
collaborative efforts. For Turkish, there are \num{275460} inflected forms of \num{3579} unique
entries (some are multiword expressions).
The data is \weblink{https://github.com/unimorph/tur}{publicly available}.

TrLex \citep{aslan2018} converts the word entries of the Turkish Language Association (TDK) dictionary
into an XML format with separate fields (e.g., lemma, POS tag, origin, meaning, example)
and annotates them with morphological segmentation for derivational suffixes. In addition, there is a
phonological representation that encodes how entries undergo Turkish morphophonemic rules.
There are \num{110960} entries in total. It is possible to obtain the version with morphological segmentation and POS tags through email communication with the authors.

Universal Derivations  \citep[UDer,][]{kyjanek2019} proposes a unified scheme for
derivational morphology. The Turkish part of the project uses EtymWordNet \citep{deMelo2010}
as a resource.
The unified resources of \num{20} languages are currently
\weblink{https://lindat.mff.cuni.cz/repository/xmlui/handle/11234/1-3236}{available online}.
In the Turkish part, there are \num{1937} unique entries and it adds up to \num{7774} derived word forms.
However, there are also errors (e.g.,  most of the derivational entries are inflectional forms).

\citet{oflazer-etal-2004} built a multiword expression extraction tool that exploits the morphological analyzer lexicon of \citet{oflazer1994} for non-lexicalized and semi-lexicalized multiwords. The lexicalized multiwords collected in this study are \weblink{https://github.com/ozlemcek/TrMWELexicon}{publicly available}.

\citet{zeyrek2019} built a lexicon of discourse connectives extracted from
Turkish discourse corpora \citep{zeyrek2013,zeyrek2017,zeyrek2018}. The lexical entries
are annotated with a canonical form, orthographic variants, corpus frequency and POS
tags. The data is part of a publicly available
\weblink{http://connective-lex.info}{multilingual connective lexicon database}.

\subsection{Morphological analyzer lexicons}\label{sec:morphological-analyzers}

Since Turkish is a morphologically rich language,
morphological analysis and lexical resources related
to morphological analyzers have been a central component of Turkish NLP.
Early attempts of building morphological analyzers
date back to \citet{koksal1975} and \citet{hankamer1986}.
The first practical and most influential morphological analyzer is by \citet{oflazer1994}.
This analyzer has been used in a large number of studies.
It is also extended by \citet{oflazer2006}
to produce pronunciations as well as the written forms.
However, these resources are developed using non-free Xerox tools,
and their availability and license is unclear.
More recently, increased availability of free finite-state tools (e.g., SFST \citep{schmid2005},
HFST \citep{linden2009} and Foma \citep{hulden2009})
resulted in a relatively large number
of freely available morphological analyzers during the last decade.
The free/open-source morphological analyzers written in conventional
finite-state tools include
\citet[implemented with Xerox languages using Foma/HFST]{coltekin2010},
\citet[implemented with SFST]{kayabacs2019},
and \citet[implemented with OpenFST]{ozturel2019}.
Another popular tool is Zemberek \citep{akin2007}
which is an open-source application written in Java
for various NLP tasks including morphological analysis.

\subsection{WordNets and other lexical networks}
\label{wordnets}

A WordNet is a lexical database where lexical items (words and phrases) are grouped into synonym sets (``synsets''). All synsets are organized in a tree structure with the hypernymy relation. Some synsets also bear additional semantic relations such as antonymy. The original WordNet for English was built at Princeton University starting in 1990 \citep{Fellbaum98} and over the years WordNets have been developed for more than \num{200} languages \citep{WNList}.

The first Turkish WordNet \citep{bilgin2004,cetinoglu2018} is developed as part of the BalkaNet project \citep{Stamou2002}, which has a direct influence on the selection of synsets. As the main goal of the project was to ensure parallelism among six Balkan WordNets as well as direct mapping to Princeton WordNet and to the eight WordNets of EuroWordNet \citep{Vossen1998}  the majority of the synset concepts are translated from Princeton WordNet. The remaining synsets are comprised of Balkan-specific concepts and frequent Turkish words. Synonyms of translated synsets and their semantic relations are populated by exploiting the TDK dictionary. The Turkish WordNet is \weblink{https://github.com/ozlemcek/twn}{publicly available}.

KeNet \citep{ehsani2018}, on the contrary, follow a bottom-up approach for creating their version of the Turkish WordNet and take the concepts in the TDK dictionary as their starting point. These concepts are semi-automatically grouped into synsets and verified manually. They also exploit Turkish Wikipedia for hypernymy relations. The resulting WordNet is standalone. This is partially improved by \citet{Bakay2019} who match \num{4417} of most frequent English senses from Princeton WordNet to KeNet synsets. KeNet is also \weblink{https://github.com/StarlangSoftware/TurkishWordNet}{publicly available}. 

Another popular lexical network is a PropBank that annotates semantic relations between predicates  and their arguments. The first example is the English PropBank \citep{palmer2005} and several PropBanks followed over the years, including Turkish ones.
The first Turkish PropBank is annotated by \citet{sahin2018} on top of the IMST dependency treebank. Later, it was adapted to the UD version of the same treebank. The annotation scheme includes numbered arguments (up to six), which correspond to the core arguments of a verb (e.g., \texttt{Buyer} is Arg0 for the predicate \textit{buy}), and \num{14} temporary roles that represent adjunct-like arguments (e.g., DIR for direction) of a verb. The resource is available by requesting it via \weblink{http://tools.nlp.itu.edu.tr/Datasets}{a license form}.

Another PropBank for Turkish is constructed by \citet{ak2018a} on top of
the constituency treebank of Turkish \citep{yildiz2014}. In this case, numbered arguments are up to four and nine temporary roles are employed. \citet{ak2018b} compare their PropBank to that of \citet{sahin2018}.
The same group has continued working on PropBanks and released TRopBank \citep{kara2020prop} which employ numbered arguments up to four and a different set of semantic role labels. While the former paper has a broken link, the latter version is publicly available \weblink{https://github.com/StarlangSoftware/}{online}.
The number of sentences that are annotated and the average of arguments
per predicate are provided in Table \ref{tbl:propbanks} for all PropBanks.

\begin{table}
  \caption{Turkish PropBanks and their basic statistics.`Avg.\ arg/prd' stands for average arguments per predicate.
  }\label{tbl:propbanks}
  \centering
  \begin{tabular}{llS[table-format=4.0,table-number-alignment=right]S[table-format=1.1,table-number-alignment=right]}
%  \begin{tabular}{llrr}
    \toprule
    PropBank     & &  \multicolumn{1}{l}{Sentences} & {Avg.\ arg/prd} \\
    \midrule
    Turkish PropBank &\citet{sahin2018}    & 5635 & 1.80 \\
    Turkish PropBank &\citet{ak2018a}    & 9560 & {--} \\
    TRopBank      &\citet{kara2020}      & {?} &  1.68 \\
%   Turkish PropBank &\citet{sahin2018}    & \num{5635} & \num{1.80} \\
%   Turkish PropBank &\citet{ak2018a}    & \num{9560} & {--} \\
%   TRopBank      &\citet{kara2020}      & ? &  \num{1.68} \\
     \bottomrule
  \end{tabular}
\end{table}

\weblink{https://github.com/commonsense/conceptnet5/}{ConceptNet} 
\citep{speer2018}
is a semantic network that creates knowledge graphs from several multilingual resources such as infoboxes of Wikipedia articles, Wiktionary, and WordNets.
The concepts are connected with intralingual and interlingual links. 
304 languages take part in the project with varying vocabulary sizes. Turkish is in the mid-range with a vocabulary size of \num{65892}.
As a follow-up project, \citet{Speer2017} have developed
\weblink{https://github.com/commonsense/conceptnet-numberbatch}{multilingual embeddings} based on ConceptNet.
Both resources are available for download.

FrameNet \citep{baker1998} is a lexical database that structures predicates and their arguments as frames. The first FrameNet is developed for English and over the years other languages have built their FrameNets. A Turkish FrameNet was recently introduced \citep{marsan2021}. It is designed to be compatible with KeNet \citep{ehsani2018,Bakay2019} and TRopBank \citep{kara2020} by using the same lemma IDs. In total there are 139 frames that include 2769 synsets, which corresponds to 4080 predicates. The FrameNet is available \weblink{https://github.com/StarlangSoftware/TurkishFrameNet}{online}.

\subsection{Word embeddings and pre-trained language models}

Word embeddings have gained substantial ground with the rise of neural models.
As a consequence, several pretrained models for Turkish were released,
as well as multilingual models.
For Turkish, there are
Word2vec \citep{sen2014,gungor2017},%
\footnote{Also at \url{https://github.com/akoksal/Turkish-Word2Vec} without an associated publication.}
\weblink{http://www.cs.cmu.edu/~afm/projects/multilingual_embeddings.html}{GloVe} \citep{ferreira2016},
\weblink{https://fasttext.cc/docs/en/crawl-vectors.html}{fastText} \citep{grave2018},
\weblink{https://github.com/HIT-SCIR/ELMoForManyLangs}{ELMo} \citep{che2018},
and several
\weblink{https://github.com/stefan-it/turkish-bert/}{BERT}
\citep{schweter2020} models available for download.
\citet{kuriyozov2020} created cross-lingual fastText embeddings
aligned to English embeddings for five Turkic languages.
The embeddings as well as the dictionaries they used
for alignments are
\weblink{https://github.com/elmurod1202/crosLingWordEmbTurk}{publicly available}.
Turkish is also part of the multilingual embeddings such as
\weblink{https://github.com/facebookresearch/MUSE}{MUSE} \citep{conneau2017word},
\weblink{https://github.com/google-research/bert}{mBERT}
\citep{devlin-etal-2019-bert},
  and \weblink{https://ai.facebook.com/blog/-xlm-r-state-of-the-art-cross-lingual-understanding-through-self-supervision/}{XLM-R} \citep{conneau2020}.

\subsection{Sentiment, emotion and other application-specific lexicons}
\label{ssec:app-specific-lexicon}

Emotion and sentiment lexicons play an important part for emotion and sentiment
analysis approaches.
\citet{cakmak2012} has created an emotion words lexicon for Turkish by translating
EMO20Q's list of English emotions \citep{kazemzadeh2011} and adding synonyms for
some translations. The total list of \num{197} words is not publicly available.
A more recent emotion lexicon is introduced by \citet{tocouglu2019tel},
which contains scores for six emotion categories across \num{4966} lexical entries.
The lexicon is
\weblink{http://demir.cs.deu.edu.tr/downloads/}{available online}
for non-commercial use.

\citet{vural2013a} has translated SentiStrength \citep{Thelwall2012} to obtain a
sentiment lexicon. SentiStrength assigns positive and negative scores to a set
of words as well as creating lists of booster words, negation words,
idioms, and emoticons. All lists are created also for Turkish. The paper does not provide information about the availability of the dataset.

\citet{chen2014} have automatically generated sentiment lexicons for \num{136} languages
including Turkish, using English as the source language. They used Wiktionary,
Google Machine Translation API, and WordNets as mapping resources. About \SI{60}{\percent} of
the words are negative in the Turkish lexicon. The dataset is accessible via
\weblink{https://sites.google.com/site/datascienceslab/projects/multilingualsentiment}{the authors' webpage}.

\citet{dehkharghani2015} utilize Turkish WordNet \citep{cetinoglu2018} to create
a sentiment lexicon named SentiTurkNet. They first manually label each synset
with positive, negative, and neutral polarity. Then they make use of
the synset mapping between Turkish and English WordNets \citep{Fellbaum98} so
that by transitivity SentiTurkNet can inherit the polarity strength scores of
SentiWordNet \citep{baccianella2010}, a sentiment lexicon which is built
on top of the English WordNet. The dataset is \weblink{http://myweb.sabanciuniv.edu/rdehkharghani/files/2016/11/SentiTurkNet.zip}{publicly available online}.

\begin{table}
  \caption{The statistics for Turkish sentiment lexicons.
  For SentiTurkNet, each synset member is counted as one token.
  }\label{tbl:sentiment}
  \centering
  \begin{tabular}{l@{\hspace{6pt}}l@{\hspace{6pt}}S[table-format=5.0]@{\hspace{6pt}}l}
    \toprule
    Sentiment Lexicon    &  & {Tokens} & Polarity \\
    \midrule
    Tr SentiStrength &\citet{vural2013a}    &  1366 & Pos (1-5), Neg (1-5) \\
    Multilingualsentiment &\citet{chen2014}     & 2500 & Pos, Neg \\
    SentiTurkNet      &\citet{dehkharghani2015}      & 21623 &  Pos (0-7),Neg (0-7),Neut \\
    \bottomrule
  \end{tabular}
\end{table}

A normalization lexicon for social media text normalization is
presented in \citet{demir2016}.
The lexicon is demonstrated to provide accurate normalization,
but statistics of the lexicon are not specified.
The paper notes that the resource is publicly available
without indicating a method for obtaining it.

\section{General Discussion}\label{sec:discussion}

The focus of our survey is exploring data sources for Turkish NLP applications, computational/quantitative linguistics research,
as well as (digital) humanities research that may benefit from linguistic data.
In this section, we list some of our observations,
followed by a short list of recommendations
for future efforts on creating language resources. 
Although we found them to be more prevalent
in comparison to efforts for resource rich, well-studied languages,
most of the observations and recommendations are not specific
to Turkish language resource creation efforts.
We believe these recommendations could particularly be useful for linguistic resource creation efforts
for languages for which there are relatively few data-driven studies,
and the conventions and traditions in the field are not yet well established.

\textbf{Availability and maintenance of resources}
Although it is not unique to Turkish resources,
we have encountered difficulties about finding and/or confirming
the availability of the data sources.
The locations of published resources are not always stable and/or permanent.
The URLs indicating the location of the resources in papers
or on the webpages of the authors or institutions are not always maintained
and the resources often disappear after publication.
Although our efforts to reach out to the authors/creators
of the resources often yielded positive results,
it is desirable to diminish these barriers to keep up with the fast-paced research community.

Another difficulty about the availability and maintenance
of the resources is related to the publication traditions
in other fields outside computational linguistics.
In particular, most papers published in general computer science venues
(e.g., in ACM conferences or journals)
do not include information about the availability of their data sources.
In some fields (e.g., speech processing),
it is more common to make the resource available for a fee
which reduces their accessibility especially for early stage researchers
or researchers with limited research budgets.
In addition,
the majority of published resources for Turkish do not include an
explicit license or ethical statement concerning
collection, distribution and use of the data.

\textbf{Awareness of earlier work}
Although it is not unique for the research papers
in Turkish Computational Linguistics,
earlier research/resources (either for Turkish or other languages)
are not cited or there is only a short list of references
ignoring other relevant research.
This results in many repetitions and inconsistencies
in the newly created resources.%
\footnote{This criticism does not refer to the creations of similar resources
 from multiple independent groups.
 As the CL and NLP become more and more data driven,
 we definitely benefit from more data,
 and well-informed and yet different approaches to the same problem.}
For example, the inconsistencies and the lack of communication
during the creation of different treebanks for Turkish
have been brought up by multiple researchers
(see Section~\ref{sec:treebanks}).

Another, related, observation is the tendency to create new resources
rather than improving the existing ones.
This leads to substantial effort put into the same work,
without clear improvements over the earlier systems.
For example,
despite the fact that some of the earlier morphological analyzers
reviewed in Section~\ref{sec:morphological-analyzers}
have been available with free licenses,
a large number of new ones were created without
a clear statement of difference or comparison.
Similar observations can be made for other resources
(e.g., WordNets) and annotation tools as well,
e.g., improving existing annotation tools could be more useful
than creating new tools which are often used in a single project.

Although most research in computational linguistics is publicly available,
there is also a need for better communication among scholars
to inform each other and collaborate on the ongoing projects, efforts and plans for building and maintaining linguistic resources. In addition, there is a need for more communication and collaboration between linguists and computational linguists for creating, annotating and analyzing language related data and resources. 

\textbf{Issues about multilingual resources}
There is a rapid increase in the efforts of building
massively multilingual resources for various tasks and applications.
We covered some of these efforts in our survey as well.
By necessity, these efforts involve either opportunistic annotations
(e.g., use of already existing information for other purposes,
like word lists in Wiktionary),
or rely heavily on crowd sourcing and/or automatic annotations.
However, a potential pitfall is the lack of quality checks
for these resources which do not necessarily involve linguistic expertise
in each language included in the resource.
For example, there are serious issues about the inflectional
and derivational lexicons discussed in Section~\ref{ssec:lexicons}.
Although these multilingual resources are useful in many tasks,
one should be aware of potential quality issues as well.

\textbf{Issues about translated resources}
Like for other languages,
automatic or manual translations of large datasets created originally
for English are also translated to Turkish.
Although this approach is interesting as it yields parallel resources,
the resource created in this manner includes effects of `translationese',
as well as additional errors that may be introduced during the translation process.
Translated datasets may even include correct translations that are
not appropriate for a particular task.
For example,
as noted by \citet{budur2020},
the inferential relation for two English sentences
may be reversed when translated to Turkish,
because Turkish pronouns are gender-neutral.
In general, the same type of inference in the original language
may not be applicable in the translation.
Similar problems are difficult to prevent
with automatic translations or non-expert human translations performed
without paying attention to the purpose of the dataset.

\textbf{Issues about quantity and quality}
With respect to the quantity of resources,
Turkish may be considered close to a  `resource-rich' language.
For example,
Turkish has the largest number of treebanks (together with English)
in the Universal Dependencies repositories (as of UD version \num{2.10}).
However, 
most Turkish treebanks are smaller in size in comparison to
treebanks in other languages,
and quality and inconsistency issues have been raised
in multiple earlier studies (see Section~\ref{sec:treebanks}
for a short discussion and pointers to relevant papers).
The same trend can be observed in other types of resources as well.
For example, \citet{aksan2018} report partial results of a questionnaire
conducted in 2011,
where Turkish NLP specialists were asked
to rate the quantity and quality of the available corpora
on a scale of \num{0} to \num{6}.
The results indicate rather low judgments,
\num{1.9} for quantity and \num{2.9} for quality.%
\footnote{The complete results of the questionnaire are not published.
  Hence, the wording of the questions, and the type of corpora
  queried are not clear.}
Although the quantity issues seem less of a problem currently,
the number of linguistic resources for Turkish are still relatively
low compared to well-studied European languages.

Overall,
it is difficult to qualify Turkish as a `low-resource language'
based on the breadth and depth of the resources available.
However,
the resources are rather scattered across different fields,
and there are issues of availability and quality.
In sum, it is probably apt to classify Turkish as a `resource poor' language
(following the terminology used by \citet{zaghouani2014} for Arabic).

\textbf{Descriptions of datasets}
A related problem in the publications introducing resources is  
the lack of sufficient descriptions.
In some cases, even the basic statistics about the data
are not presented or it is difficult to interpret
the statistics due to unclear units of measurements.
There is also a need for better descriptions of
proper quality assurance procedures,
metrics and inter-annotator agreements (IAA).
Lack of proper linguistic glosses and translations
in the provided examples also create extra barriers
for readers without any Turkish background to understand
and evaluate the research article and/or the data resource.

\textbf{Gaps in the existing resources}
Although there are a number of sources for (social media) text normalization,
we are not aware of any publications on datasets of spelling or grammar errors.%
\footnote{\weblink{https://extensions.libreoffice.org/en/extensions/show/20565}{A new spelling dictionary} with an associated tool
has been announced during the final revisions of this paper.}
Similarly, there is no known learner corpus or resources that can help
second language research and practice for Turkish.

Another general area with no or little resources is semantics.
Except for the lexical resources listed in Section~\ref{sec:lexical},
we are not aware of any semantically annotated corpora
(e.g., one that would be used for semantic parsing).
There is also a lack of benchmark datasets for assessing
pre-trained word or text representations
(word embeddings, or pre-trained language models).
So far, most linguistic resources available for Turkish
aim to be domain independent.
If a resource is domain-specific,
it is often due to practical reasons rather
than a specific interest in this particular domain.
On the other hand,
domain-specific data is crucial for NLP applications.
Although the uses of unpublished datasets were reported
in earlier literature
\citep[e.g., a corpus of radiology reports by][]{hadimli2011},
there is a big gap in domain-specific datasets for
critical domains or sub-fields like biomedical, legal or financial NLP.

There is also a need for more systematic data collection
and analysis of dialectal and sociolinguistic variation
with easy-to-access language resources \citep{dogruoz}.

\textbf{A concise list of recommendations}
The issues raised above in this section have some
rather obvious solutions.
Nevertheless, the concise list below may be
beneficial for future resource creation efforts.

\begin{itemize}
  \item \emph{Publish your corpora,
    and publish it on permanent (or long-lasting) venues.}
    Beyond the value of the published data and code for reproducibility,
    published data allows others to study the data in ways 
    creators of the data cannot possibly foresee.
    Furthermore, growing evidence suggests that the papers
    that publish their data get more recognition
    \citep{wieling2018,colavizza2020}.
    It is also important to publish the data in locations that would not
    disappear shortly after the publication.
    Our experience in this survey shows that the data shared
    through personal and also institutional webpages
    often become inaccessible as authors move to other institutions,
    or their research interests change.
    As a result, publishing the data in general repositories like
    \weblink{https://zenodo.org/}{Zenodo}
    and \weblink{https://osf.io/}{OSF},
    or \weblink{https://www.clarin.eu/}{CLARIN} repositories
    that are more specialized for language resources is a better choice
    than personal and institutional webpages.
    Similarly, to our experience,
    software development infrastructures
    like \weblink{https://github.com}{GitHub} also provide
    stable locations for publishing linguistic data.

  \item \emph{Describe all aspects of the corpora adequately.}
    As we occasionally noted above, a large number of papers we reviewed
    do not describe the resources introduced sufficiently.
    It is important for a paper to include information
    on aspects of the corpora such as,
    size, label distribution, source material, sampling method,
    as well as indications of annotation quality (e.g., IAA)
    in proper units and using proper metrics for the task at hand.
    Being aware of the earlier recommendations
    \citep[e.g.,][]{ide2017,bender2018,gebru2020}
    for resource creation efforts and their descriptions
    would be useful for any annotation or curation project.

  \item \emph{Be explicit about the licensing and potential ethical issues.}
    Although major computational linguistics venues started to require
    statements about legal and ethical aspects of data collection and sharing,
    not all the venues require such statements.
    It is important to be aware of the existing guidelines,
    such as ACM code of ethics \parencite{acm-guidelines},
    or the guidelines adapted by major CL conferences,%
    \footnote{For example, NAACL guidelines
      at \url{https://2021.naacl.org/ethics/faq/} which is
      also adapted by some of the other major CL conferences.}
    as well as the recent discussion in the field
    \parencite[e.g.,][]{suster2017,rogers2021}.
    Even though the common guidelines may not fit every task,
    or every legal jurisdiction, being aware of potential issues,
    and being explicit about the legal and ethical considerations
    during data collection and annotation is important.
    The lack of clarity around these issues may also reduce the usability of
    the data (and hence, the recognition the creators may receive).

  \item \emph{Before creating a new resource,
    perform a thorough literature review of the relevant research,
    consider improving existing resources,
    and collaborating with other scholars in the field.}
    As evidenced by the lack of citations in published papers,
    most resources are built from scratch,
    not paying attention to the lessons learned in the earlier work.
    The quality of linguistic resources could be improved by awareness
    of earlier work and more collaboration between different groups.
    Besides individual efforts from researchers and reviewers,
    a regular meeting of CL/NLP researchers and practitioners
    working on Turkish (and possibly Turkic languages)
    may help alleviate this problem.
    Although a number of `first attempts' were made for such meetings,
    unlike many other CL communities,
    no regular/stable meeting has been established so far.

  \item \emph{Contribute to multilingual resource creation efforts.}
    One of the issues we observed above with large-scale, multilingual 
    resources is the lack of quality in Turkish data in these efforts.
    Bringing the language expertise of Turkish (computational) linguists
    in these projects would definitely improve the quality of these efforts,
    which, in turn, would be beneficial to the CL/NLP studies in Turkish.
\end{itemize}

\section{Conclusion}

Our goal in this survey was to present
a comprehensive summary of language resources NLP
and computational/quantitative linguistic research for Turkish.
In addition to the resources listed in our survey,
we also provide a companion website (\url{https://turkishnlp.github.io})
which includes links to even more Turkish resources,
and we will update it regularly.
In this way, our survey and the companion website will serve as
stable and sustainable resources for researchers across disciplines
(e.g., linguistics, NLP) who are currently working on Turkish.
In addition, researchers who are not currently working on Turkish
but who need linguistic resources outside their current expertise
and/or those who are interested in including Turkish
in multi- or cross-lingual tasks could benefit from our contribution as well. 

Besides the comprehensive overview of the resources,
we have also summarized some of the common problematic issues
and gaps in the field and provided a set of short suggestions
for future resource creation efforts.
We cautiously note that not all the problematic issues
could easily be resolved by individual researchers
and research groups immediately.
Some of these issues require long-term collaborative efforts
within the community as well as substantial support
from academic funding agencies for further research. 
The issues we raise in this paper are based on our impression
from published papers and cursory inspection of the available corpora.
To understand the factors behind these issues better
and propose informed solutions,
future studies with in-depth analyses
(e.g., through questionnaires directed to creators and users of the resources,
or more systematic inspection of the available data) can be helpful.
Similarly, effectiveness of the guidelines
(offered in papers we cite in Section~\ref{sec:discussion})
may also be measured in future experimental studies.

In short, we hope that our survey and its companion webpage will serve
as a useful reference for locating resources
for existing fundamental and applied research and
for creating future resources and projects for Turkish and/or other languages. 

\printbibliography

\end{document}